\documentclass{article}

    \PassOptionsToPackage{numbers, compress}{natbib}

    \usepackage[preprint]{neurips_2020}

\usepackage[utf8]{inputenc} 
\usepackage[T1]{fontenc}    
\usepackage{hyperref}       
\usepackage{url}            
\usepackage{booktabs}       
\usepackage{amsfonts}       
\usepackage{nicefrac}       
\usepackage{microtype}      %
\usepackage{float}

\usepackage{placeins}
\usepackage{times}
\usepackage{epsfig}
\usepackage{graphicx}
\usepackage{amsmath}
\usepackage{amssymb}
\usepackage{subfig}
\usepackage{multirow}
\usepackage{array}
\usepackage{xcolor}
\usepackage{booktabs}
\usepackage{dblfloatfix}
\def \hfillx {\hspace*{-\textwidth} \hfill}

\usepackage{bm}
\title{Improving Multimodal Accuracy Through Modality Pre-training and Attention}

\author{%
  Aya Abdelsalam Ismail\textsuperscript{1},  \space Mahmudul Hasan\textsuperscript{2},  \space Faisal Ishtiaq\textsuperscript{2}  \\
    \texttt{asalam@cs.umd.edu,  mahmud.ucr@gmail.com, Faisal\_Ishtiaq@comcast.com} \\
\textsuperscript{1}Department of Computer Science, University of Maryland\\ \textsuperscript{2}Comcast Labs, Washington, DC, USA
}

\begin{document}

\maketitle

\begin{abstract}
Training a multimodal network is challenging and it requires complex architectures to achieve reasonable performance. We show that one reason for this phenomena is the difference between the convergence rate of various modalities. We address this by pre-training modality-specific sub-networks in multimodal architectures independently before end-to-end training of the entire network. Furthermore, we show that the addition of an attention mechanism between sub-networks after pre-training helps identify the most important modality during ambiguous scenarios boosting the performance. We demonstrate that by performing these two tricks a simple network can achieve similar performance to a complicated architecture that is significantly more expensive to train on multiple tasks including sentiment analysis, emotion recognition, and speaker trait recognition.

\end{abstract}

\section{Introduction}


Multimodal learning \cite{ngiam2011multimodal} means using data from different modalities such as acoustic, visual and language information to perform a certain task. It involves understanding the role each modality plays in the task (intra-modal dynamics) and how modalities interact with each other (inter-modality dynamics). Previous work on multimodal learning focuses on increasing model complexity to boost accuracy. 
Such sophisticated architectures are needed to overcome the difficulty of training multimodal networks \cite{wang2020makes,kiela2019supervised}. We believe, 
the main reason behind this training difficulty is that various modalities tend to converge and generalize at different rates (shown in the first column of Figure \ref{fig:converagnce}).  Training multimodal architecture end-to-end with randomly initialized sub-networks produces a sub-optimal solution.


We address the shortcomings above by taking a two-step approach while training the multimodal network: (1) Each sub-network is pre-trained independently to learn intra-modal dynamics (allowing each modality to convergence at its own rate). (2) Different modalities are then fused by an attention mechanism (that decides which modalities are most important in a particular example at test time) and the entire network is then fine-tuned end-to-end. The entire architecture is shown in Figure \ref{fig:Model}

 We empirically show that training of multimodal architecture end-to-end from scratch will bias the network towards the modality that has a faster convergence rate. We then show that adding the pre-training step sufficiently improves the performance of multimodal networks. We evaluate our proposed training methods on three publicly available datasets for three multimodal tasks. Our model with simple modality-specific sub-networks shows competitive performance when compared to other models on all three tasks.
 
  

\section{Modality Attention Network with Two Step Training}
\begin{figure}[h]
\centering
\includegraphics[width=\textwidth]{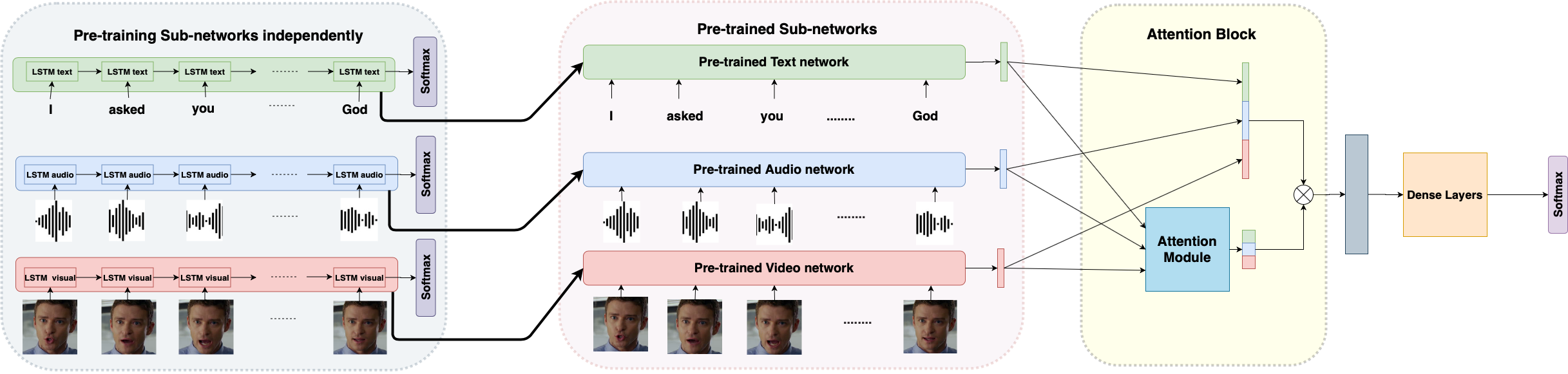} 
\caption{An overview of the proposed Modality Aware Attention Network. MAN contains two main components: (1) Independently pre-trained sub-networks that learn the representation of different modalities. (2) An attention module that decides which modality is important in a particular example.}
\label{fig:Model}
\end{figure}

Modality Attention Network (\textbf{MAN}) has two main components: (1) \textbf{Sub-networks} - these are independently trained bidirectional LSTM for each modality such as text, audio, and video for emotion recognition task. (2) \textbf{Attention Block} - this block estimates a relative salience score for each modality. MAN overview is shown in Figure \ref{fig:Model}.

\textbf{Sub-networks:}
The goal of sub-networks is to learn intra-modal dynamics and produce an embedding for each modality that will be used by the attention block. We use a bidirectional Long-short Term Memory (LSTM) \cite{schuster1997bidirectional} for each sub-network. LSTMs have been widely used in modeling temporal data
in many tasks including text classification  \cite{sutskever2014sequence}, video classification \cite{donahue2015long}, and voice activity detection \cite{hughes2013recurrent}.
Note that LSTM can be replaced with any neural architecture and the sub-networks are not required to have the same architecture.

\textbf{Modality-Specific Pre-training:}
The modality-specific sub-networks are first pre-trained independently on the task of interest. This is done for two reasons: (1) Pre-training allows better learning of intra-modal dynamics since the learning process is not affected by any other modalities at this time. (2) Data from different modalities have different convergence rates, if we train the network end-to-end without the pre-training step, the network tends to focus on the modality that converges faster ignoring others.

The softmax layer at the end of the modality-specific sub-network is removed after pre-training and the last hidden layer is used as a feature representation of that modality and is given as the input to the attention block as shown in Figure \ref{fig:Model}.

Formally, let the input to the language sub-network be denoted as $L=\left[l_0,\dots,l_t,\dots, ,l_n\right]$ where $l_t$ is the $t ^\text{th}$ word in the sentence and $l$ is the feature embedding. Similarly, the input to acoustic and visual sub-network is denoted as $A$ and $V$ respectively. The output of the sub-networks is the last hidden state of modality-specific network can be written as:
$ h_l=\text{Net}_l(L), \;  \;  h_a=\text{Net}_a(A), \; \;  h_v=\text{Net}_v(V)$.

\textbf{Attention Block:}
Hidden layers from sub-networks $h_l,h_a,$ and $h_v$ each have a same dimension of $N$. Matrix $H$ is the input to attention block where $H=[h_l,h_a,h_v]$; $H$ has a dimension of $m\times N$, where $m$ is the number of modalities. The output of the attention block is a weight matrix $A$.

Attention matrix $A$ is given as:
$ softmax \left(W_{2} \left(\tanh \left(W_{1}H^{T}  + b_1\right)\right) + b_2\right)
$. Weight matrix $W_{1}$ and $W_{2}$ have dimensions of $ k \times N$ and $1 \times k$ respectively, where k is a hyper-parameter; $b_1$ and $b_2$ are bias parameter and the $softmax()$ ensures that all the computed weights are sum to 1. In the attention matrix $A=\left[w_l,w_a,w_v\right]$, weights $w_l,w_a,$ and $w_v$ represent importance of different modalities. 

\textbf{Output of MAN:}
The hidden layer of each sub-network is multiplied by its weights, which is then concatenated together and passed to a fully connected layer and a $softmax$ layer for the final classification. The entire architecture is then trained end-to-end together, which fine-tunes the sub-networks in the process. This helps to capture both intra-modal and inter-modal dynamics.

\section{Experiments}
We evaluated the proposed framework on three real-world applications: sentiment analysis, emotion recognition, and speaker trait recognition. The experiments have been designed to investigate the following: (a) The performance of MAN in terms of accuracy, complexity, and training time when compared to other networks. (b) The effect of pre-training the sub-networks versus end-to-end training of the entire network for multimodal networks. (c) The importance of attention module in differentiating between important and ambiguous modalities. 
Detailed description of datasets, baselines and experiments is available in the supplementary material.

\textbf{Datasets:} We selected three publicly available multimodal datasets that contain spoken language, acoustic and visual information. Multimodal Sentiment Analysis 
 \textit{CMU MOSI} dataset \cite{zadeh2016multimodal}, Multimodal Emotion Recognition  \textit{CMU MOSEI} dataset \cite{zadeh2018multimodal} and Multimodal Speaker Trait Recognition  Persuasion Opinion Multimodal \textit{POM} dataset \cite{park2014computational}.
 
\textbf{Computational Descriptors:}
\textit{Language Features:} GloVe \cite{pennington2014glove} word embeddings are used to convert the transcripts into word embeddings.  \textit{Acoustic Features:} COVAREP  \cite{degottex2014covarep} acoustic analysis framework is used to extract low level acoustic features. \textit{Visual Features:} Facial expression analysis toolkit FACET \cite{iMotion}  is used as visual feature extractor.

\textbf{Baseline Models} 
We compare the performance of MAN to a variety of  models for multimodal language analysis; all models are trained using the same feature embeddings:
\textit{\textbf{LF-LSTM}} Late Fusion LSTM,
\textit{\textbf{MAF}} Modality Attention Fusion \cite{gu2018hybrid}, \textit{\textbf{TFN} } Tensor Fusion Network \cite{zadeh2017tensor}, \textit{\textbf{LMF}} Low-rank Multimodal Fusion \cite{liu2018efficient} ,  \textit{\textbf{MFN}} Memory Fusion Network\cite{zadeh2018memory},
  \textit{ \textbf{MFM}} Multimodal Factorization Model \cite{tsai2018learning} and 
 \textit{\textbf{RAVEN}} Recurrent Attended Variation Embedding Network \cite{wang2019words}.

\textbf{Evaluation Metrics}
For regression, we report Mean Absolute Error. For classification accuracy, we report $A^C$ where $C$ is the number of classes. In addition, we report the mean epoch training time in seconds and  inference time in milliseconds.

\section{Results and Discussion}

\paragraph{Performance of MAN}

We show the results in Table \ref{tab:Results}. There is no single model that always outperform others in all metrics across different tasks, in terms of accuracy and MAE. MAN shows competitive performance when compared against the other models. Generally best performing models are RAVEN and MAN; while  MAN requires less training time.

\begin{table*}[hbtp!]
  \centering
    \resizebox{\textwidth}{!}{
    \begin{tabular}{l|cccc|cccc|cccc}
        \toprule
           \multicolumn{1}{c|}{}     & \multicolumn{4}{c|}{ \textbf{CMU-MOSI}} & \multicolumn{4}{c|}{\textbf{CMU-MOSEI}} & \multicolumn{4}{c}{\textbf{POM}} \\

         Method &$A^2$   & \multicolumn{1}{c}{MAE}    &\multicolumn{1}{c}{ep.} &\multicolumn{1}{c|}{In.}   & \multicolumn{1}{c}{$A^4$}     & \multicolumn{1}{c}{MAE}     &\multicolumn{1}{c}{ep.}
         &\multicolumn{1}{c|}{In.} &
         \multicolumn{1}{c}{$A^7$}    & \multicolumn{1}{c}{MAE}   &\multicolumn{1}{c}{ep.}
         &\multicolumn{1}{c}{In.} \\
    \hline \hline
        LF-LSTM & 0.747 & 1.025 &   1.92 &0.36 & 0.645 & 0.155 & 19.02&0.04& 0.366 & 0.832 &  8.25 & 0.25\\
        MAF   & 0.762 &  1.002 & 1.19 &0.17  & 0.673 & 0.159 & 13.96 & 0.02 &  \textbf{0.381} & $0.822^*$ & 4.27 & 0.13 \\
        TFN   & 0.753 & 1.011 &1.07 &0.23 &0.701 & 0.176 & 9.72 & 0.04& 0.367 & 0.826 & 2.75 & 0.06\\
        LMF   & 0.766 & 1.002 &   0.96& 0.24 & \textbf{0.761} & 0.160  & 9.53&0.04 & 0.362 & 0.844 & 2.03 & 0.07 \\
        MFN   & 0.750 & 1.027 & 6.75 &1.16 & 0.723 & 0.163 &  79.28  &0.16 & 0.360  & 0.838 & 82.8 &1.40\\
        MFM   & 0.761 & 1.010 &  9.52 &1.52& 0.632 & 0.163 & 114.42 &0.19 & 0.366 & 0.854 &131.79 &1.94 \\
        RAVEN & $0.774^*$ & \textbf{0.960} & 20.43&3.63 & 0.536 & $0.153^*$ & 89.00& 0.61&  0.366 & $0.825$ &60.04&2.01 \\
    \midrule
    MAN   & \textbf{0.784} & $0.970^*$ &  1.76&0.37& $0.7604^*$   & \textbf{0.147} &  19.47&0.04 &$0.378 ^*$ & \textbf{0.807} &11.39 &0.28 \\
    \bottomrule
    \end{tabular}}
     
  \caption{ Results on different datasets. \textbf{$A^C$} is accuracy, \textbf{MAE} is Mean absolute error, \textbf{ep.} denotes the mean epoch training time in seconds, and \textbf{In.} is mean inference time in milliseconds. Best results are bolded and the $2^{\text{nd}}$  best is noted with $^*$.}
 \label{tab:Results}%
\end{table*}

\begin{figure*}[htb!]
    \centering
    \includegraphics[width=0.57\linewidth]{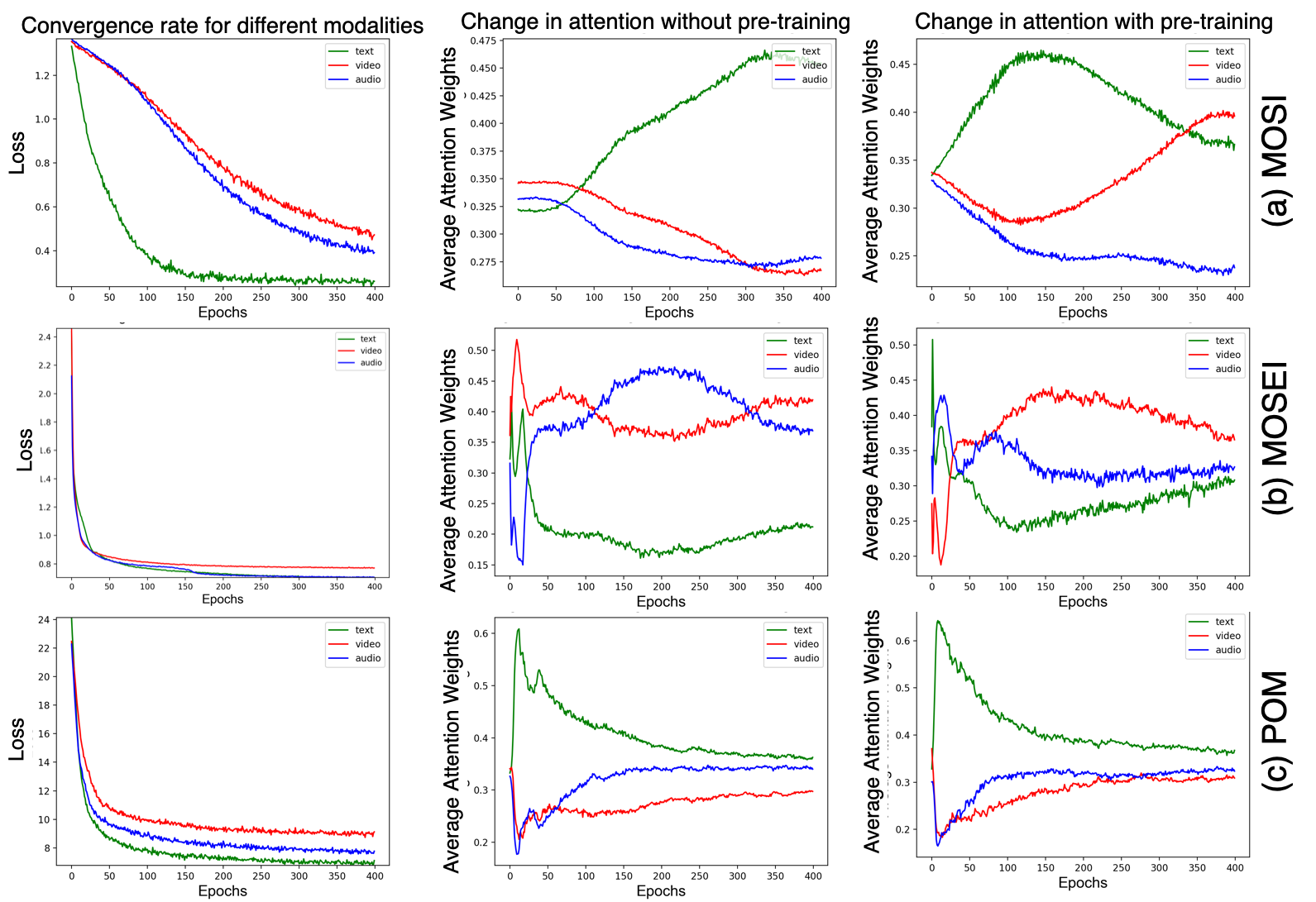}
    
        \caption{The effect of pre-training multimodal networks on attention weights. The $1^{\text{st}}$ column shows the difference in converge rate of each modalities. The $2^{\text{nd}}$  column shows the average attention weights over all examples per epochs. The $3^{\text{rd}}$ column shows the the average attention weights over all examples per epochs if sub-networks we pre-trained before added to the end-to-end network.}%
    \label{fig:converagnce}%
\end{figure*}

\textbf{Why do we need modality-specific pre-training?} \space
Different modalities converge and generalize at different rates. Consider Figure \ref{fig:converagnce}  (a), the $1^{\text{st}}$ column shows how language modality converges faster than the acoustic and the visual modality for MOSI dataset. In this case, end-to-end training leads to favoring language modality as shown in the  $2^{\text{nd}}$ column, where the attention module assigns higher weights to the language. As the network continues to learn, the effect of acoustic and visual modalities tend to diminish over time. If we pre-train the sub-networks and then fine-tune the entire architecture, we eliminate the bias introduced by the difference in convergence rate of different modalities, as shown in the  $3^{\text{rd}}$ column; where network learns that both language and visual modalities are equally important in this task. For MOSEI and POM (the  $1^{\text{st}}$ column of Figures \ref{fig:converagnce}(b) and \ref{fig:converagnce}(c) respectively), the differences in convergence rate among the modalities are not as significant as the MOSI dataset. However, training without pre-training the sub-networks leads to a biased distribution of the weights among the modalities as shown in the $2^{\text{nd}}$ column. This is fixed when we pre-train the sub-networks as shown in the $3^{\text{rd}}$ column, which eventually leads to better performance.

\textbf{The effect of pre-training on different multimodal architectures:} We repeated MOSI experiment with pre-training for \textbf{LF-LSTM} and \textbf{MAF} since they allow separation of sub-networks.
Table \ref{tab:pretraininAcc} shows that pre-training sub-networks before end-to-end training improves the accuracy on both.
\begin{table}[htbp!]
    \begin{minipage}{.4\linewidth}
       \resizebox{\textwidth}{!}{ \begin{tabular}{l|c|c}
   \multicolumn{1}{l|}{ }  &\multicolumn{1}{c|}{  LF-LSTM } & \multicolumn{1}{c}{MAF}\\
    \hline
        \hline
 w/o pre-training  &  0.747  &   0.762   \\
   + pre-training  & \textbf{0.762} &  \textbf{0.769}  \\
       \hline
    \end{tabular}}
    \caption{$A^2$ with \& without Pre-training.}
      \label{tab:pretraininAcc}%
    \end{minipage}%
    \hfillx
    \begin{minipage}{.45\linewidth}
      \centering
 \resizebox{\textwidth}{!}{
    \begin{tabular}{l|c|c}
   \multicolumn{1}{l|}{}  &\multicolumn{1}{c|}{GloVe} &\multicolumn{1}{c}{BERT}  \\
    \hline    \hline
    Text Only &0.756&0.793\\
    MAN - Atten.  &   0.762 &   0.808\\
    MAN - pre-training  &  0.765 &   0.802\\
    \hline
    MAN & \textbf{0.784}& \textbf{0.812} \\
       \hline
    \end{tabular}}
       \caption{$A^2$ on different word embeddings.}
      \label{tab:AttenAcc}%
    \end{minipage} 
\end{table}
\vspace{-0.6cm}

\begin{figure*}[hbtp!]
\centering
\includegraphics[width=0.88\textwidth]{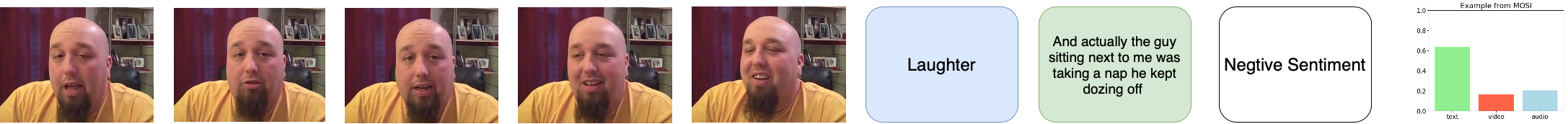} %
\caption{ MAN correctly identifies language as the most important modality.}
\label{fig:interpretation}
\end{figure*}

\textbf{Why is Attention needed?} \space
The attention module produces a single weight for each modality
pointing out which modality is important in a particular example allowing the network to focus on the feature representation of that modality which results in a boost in accuracy as shown in Table \ref{tab:AttenAcc}. In addition, attention weights can also be used as a method of model interpretation \cite{xu2015show,rocktaschel2015reasoning,li2016understanding,mullenbach2018explainable,thorne2019generating}. This in not the case in many methods such as tensor factorization, 
features from different modalities are multiplied together so it becomes unclear which modality is the most important. 
In Figure \ref{fig:interpretation}, shows an example from MOSI dataset on sentiment analysis task, where the speaker appears smiling and laughing; however, the text is negative and model was able to attend on the correct modality. More examples are available in the supplementary material. 
\vspace{-0.3cm}

\textbf{Better word embeddings better accuracy} Table \ref{tab:AttenAcc} show the difference in accuracy on MOSI when replacing the text sub-network with BERT \cite{devlin2018bert}. Since BERT produces better word embeddings than GloVe replacing text sub-network with BERT improves accuracy. Similar, to GloVe, the addition of pre-training step along with the attention module before end-to-end training improves the accuracy.

\section{Conclusion}
In this paper, we identify the difference in the convergence rate of various modalities as one of the reasons behind the difficulty of training multimodal networks. We find that this can be overcome by pre-training individual modality-specific sub-networks followed by end-to-end fine-tuning of the entire network. Finally, we can improve the accuracy further through an attention module that attends to different modalities after pre-training.

{\small
\bibliography{neurips_2020.bib}{}

\begin{thebibliography}{46}
\providecommand{\natexlab}[1]{#1}
\providecommand{\url}[1]{\texttt{#1}}
\expandafter\ifx\csname urlstyle\endcsname\relax
  \providecommand{\doi}[1]{doi: #1}\else
  \providecommand{\doi}{doi: \begingroup \urlstyle{rm}\Url}\fi

\bibitem[Ngiam et~al.(2011)Ngiam, Khosla, Kim, Nam, Lee, and
  Ng]{ngiam2011multimodal}
Jiquan Ngiam, Aditya Khosla, Mingyu Kim, Juhan Nam, Honglak Lee, and Andrew~Y
  Ng.
\newblock Multimodal deep learning.
\newblock In \emph{Proceedings of the 28th international conference on machine
  learning (ICML-11)}, pages 689--696, 2011.

\bibitem[Wang et~al.(2020)Wang, Tran, and Feiszli]{wang2020makes}
Weiyao Wang, Du~Tran, and Matt Feiszli.
\newblock What makes training multi-modal classification networks hard?
\newblock In \emph{Proceedings of the IEEE/CVF Conference on Computer Vision
  and Pattern Recognition}, pages 12695--12705, 2020.

\bibitem[Kiela et~al.(2019)Kiela, Bhooshan, Firooz, and
  Testuggine]{kiela2019supervised}
Douwe Kiela, Suvrat Bhooshan, Hamed Firooz, and Davide Testuggine.
\newblock Supervised multimodal bitransformers for classifying images and text.
\newblock \emph{arXiv preprint arXiv:1909.02950}, 2019.

\bibitem[Schuster and Paliwal(1997)]{schuster1997bidirectional}
Mike Schuster and Kuldip~K Paliwal.
\newblock Bidirectional recurrent neural networks.
\newblock \emph{IEEE Transactions on Signal Processing}, 45\penalty0
  (11):\penalty0 2673--2681, 1997.

\bibitem[Sutskever et~al.(2014)Sutskever, Vinyals, and
  Le]{sutskever2014sequence}
Ilya Sutskever, Oriol Vinyals, and Quoc~V Le.
\newblock Sequence to sequence learning with neural networks.
\newblock In \emph{Advances in neural information processing systems}, pages
  3104--3112, 2014.

\bibitem[Donahue et~al.(2015)Donahue, Anne~Hendricks, Guadarrama, Rohrbach,
  Venugopalan, Saenko, and Darrell]{donahue2015long}
Jeffrey Donahue, Lisa Anne~Hendricks, Sergio Guadarrama, Marcus Rohrbach,
  Subhashini Venugopalan, Kate Saenko, and Trevor Darrell.
\newblock Long-term recurrent convolutional networks for visual recognition and
  description.
\newblock In \emph{Proceedings of the IEEE conference on computer vision and
  pattern recognition}, pages 2625--2634, 2015.

\bibitem[Hughes and Mierle(2013)]{hughes2013recurrent}
Thad Hughes and Keir Mierle.
\newblock Recurrent neural networks for voice activity detection.
\newblock In \emph{2013 IEEE International Conference on Acoustics, Speech and
  Signal Processing}, pages 7378--7382. IEEE, 2013.

\bibitem[Zadeh et~al.(2016)Zadeh, Zellers, Pincus, and
  Morency]{zadeh2016multimodal}
Amir Zadeh, Rowan Zellers, Eli Pincus, and Louis-Philippe Morency.
\newblock Multimodal sentiment intensity analysis in videos: Facial gestures
  and verbal messages.
\newblock \emph{IEEE Intelligent Systems}, 31\penalty0 (6):\penalty0 82--88,
  2016.

\bibitem[Zadeh et~al.(2018{\natexlab{a}})Zadeh, Liang, Poria, Cambria, and
  Morency]{zadeh2018multimodal}
AmirAli~Bagher Zadeh, Paul~Pu Liang, Soujanya Poria, Erik Cambria, and
  Louis-Philippe Morency.
\newblock Multimodal language analysis in the wild: Cmu-mosei dataset and
  interpretable dynamic fusion graph.
\newblock In \emph{Proceedings of the 56th Annual Meeting of the Association
  for Computational Linguistics (Volume 1: Long Papers)}, pages 2236--2246,
  2018{\natexlab{a}}.

\bibitem[Park et~al.(2014)Park, Shim, Chatterjee, Sagae, and
  Morency]{park2014computational}
Sunghyun Park, Han~Suk Shim, Moitreya Chatterjee, Kenji Sagae, and
  Louis-Philippe Morency.
\newblock Computational analysis of persuasiveness in social multimedia: A
  novel dataset and multimodal prediction approach.
\newblock In \emph{Proceedings of the 16th International Conference on
  Multimodal Interaction}, pages 50--57. ACM, 2014.

\bibitem[Pennington et~al.(2014)Pennington, Socher, and
  Manning]{pennington2014glove}
Jeffrey Pennington, Richard Socher, and Christopher Manning.
\newblock Glove: Global vectors for word representation.
\newblock In \emph{Proceedings of the 2014 conference on empirical methods in
  natural language processing (EMNLP)}, pages 1532--1543, 2014.

\bibitem[Degottex et~al.(2014)Degottex, Kane, Drugman, Raitio, and
  Scherer]{degottex2014covarep}
Gilles Degottex, John Kane, Thomas Drugman, Tuomo Raitio, and Stefan Scherer.
\newblock Covarep—a collaborative voice analysis repository for speech
  technologies.
\newblock In \emph{2014 ieee international conference on acoustics, speech and
  signal processing (icassp)}, pages 960--964. IEEE, 2014.

\bibitem[iMo(2017)]{iMotion}
\emph{iMotion}, 2017.
\newblock URL \url{https://imotions.com}.

\bibitem[Gu et~al.(2018{\natexlab{a}})Gu, Yang, Fu, Chen, Li, and
  Marsic]{gu2018hybrid}
Yue Gu, Kangning Yang, Shiyu Fu, Shuhong Chen, Xinyu Li, and Ivan Marsic.
\newblock Hybrid attention based multimodal network for spoken language
  classification.
\newblock In \emph{Proceedings of the 27th International Conference on
  Computational Linguistics}, pages 2379--2390, 2018{\natexlab{a}}.

\bibitem[Zadeh et~al.(2017)Zadeh, Chen, Poria, Cambria, and
  Morency]{zadeh2017tensor}
Amir Zadeh, Minghai Chen, Soujanya Poria, Erik Cambria, and Louis-Philippe
  Morency.
\newblock Tensor fusion network for multimodal sentiment analysis.
\newblock \emph{arXiv preprint arXiv:1707.07250}, 2017.

\bibitem[Liu et~al.(2018)Liu, Shen, Lakshminarasimhan, Liang, Zadeh, and
  Morency]{liu2018efficient}
Zhun Liu, Ying Shen, Varun~Bharadhwaj Lakshminarasimhan, Paul~Pu Liang, Amir
  Zadeh, and Louis-Philippe Morency.
\newblock Efficient low-rank multimodal fusion with modality-specific factors.
\newblock \emph{arXiv preprint arXiv:1806.00064}, 2018.

\bibitem[Zadeh et~al.(2018{\natexlab{b}})Zadeh, Liang, Mazumder, Poria,
  Cambria, and Morency]{zadeh2018memory}
Amir Zadeh, Paul~Pu Liang, Navonil Mazumder, Soujanya Poria, Erik Cambria, and
  Louis-Philippe Morency.
\newblock Memory fusion network for multi-view sequential learning.
\newblock In \emph{Thirty-Second AAAI Conference on Artificial Intelligence},
  2018{\natexlab{b}}.

\bibitem[Tsai et~al.(2018)Tsai, Liang, Zadeh, Morency, and
  Salakhutdinov]{tsai2018learning}
Yao-Hung~Hubert Tsai, Paul~Pu Liang, Amir Zadeh, Louis-Philippe Morency, and
  Ruslan Salakhutdinov.
\newblock Learning factorized multimodal representations.
\newblock \emph{arXiv preprint arXiv:1806.06176}, 2018.

\bibitem[Wang et~al.(2019)Wang, Shen, Liu, Liang, Zadeh, and
  Morency]{wang2019words}
Yansen Wang, Ying Shen, Zhun Liu, Paul~Pu Liang, Amir Zadeh, and Louis-Philippe
  Morency.
\newblock Words can shift: Dynamically adjusting word representations using
  nonverbal behaviors.
\newblock In \emph{Proceedings of the AAAI Conference on Artificial
  Intelligence}, volume~33, pages 7216--7223, 2019.

\bibitem[Xu et~al.(2015)Xu, Ba, Kiros, Cho, Courville, Salakhudinov, Zemel, and
  Bengio]{xu2015show}
Kelvin Xu, Jimmy Ba, Ryan Kiros, Kyunghyun Cho, Aaron Courville, Ruslan
  Salakhudinov, Rich Zemel, and Yoshua Bengio.
\newblock Show, attend and tell: Neural image caption generation with visual
  attention.
\newblock In \emph{International conference on machine learning}, pages
  2048--2057, 2015.

\bibitem[Rockt{\"a}schel et~al.(2015)Rockt{\"a}schel, Grefenstette, Hermann,
  Ko{\v{c}}isk{\`y}, and Blunsom]{rocktaschel2015reasoning}
Tim Rockt{\"a}schel, Edward Grefenstette, Karl~Moritz Hermann, Tom{\'a}{\v{s}}
  Ko{\v{c}}isk{\`y}, and Phil Blunsom.
\newblock Reasoning about entailment with neural attention.
\newblock \emph{arXiv preprint arXiv:1509.06664}, 2015.

\bibitem[Li et~al.(2016)Li, Monroe, and Jurafsky]{li2016understanding}
Jiwei Li, Will Monroe, and Dan Jurafsky.
\newblock Understanding neural networks through representation erasure.
\newblock \emph{arXiv preprint arXiv:1612.08220}, 2016.

\bibitem[Mullenbach et~al.(2018)Mullenbach, Wiegreffe, Duke, Sun, and
  Eisenstein]{mullenbach2018explainable}
James Mullenbach, Sarah Wiegreffe, Jon Duke, Jimeng Sun, and Jacob Eisenstein.
\newblock Explainable prediction of medical codes from clinical text.
\newblock \emph{arXiv preprint arXiv:1802.05695}, 2018.

\bibitem[Thorne et~al.(2019)Thorne, Vlachos, Christodoulopoulos, and
  Mittal]{thorne2019generating}
James Thorne, Andreas Vlachos, Christos Christodoulopoulos, and Arpit Mittal.
\newblock Generating token-level explanations for natural language inference.
\newblock \emph{arXiv preprint arXiv:1904.10717}, 2019.

\bibitem[Devlin et~al.(2018)Devlin, Chang, Lee, and Toutanova]{devlin2018bert}
Jacob Devlin, Ming-Wei Chang, Kenton Lee, and Kristina Toutanova.
\newblock Bert: Pre-training of deep bidirectional transformers for language
  understanding.
\newblock \emph{arXiv preprint arXiv:1810.04805}, 2018.

\bibitem[Grimm et~al.(2008)Grimm, Kroschel, and Narayanan]{grimm2008vera}
Michael Grimm, Kristian Kroschel, and Shrikanth Narayanan.
\newblock The vera am mittag german audio-visual emotional speech database.
\newblock In \emph{2008 IEEE international conference on multimedia and expo},
  pages 865--868. IEEE, 2008.

\bibitem[Dhall et~al.(2012)Dhall, Goecke, Lucey, Gedeon,
  et~al.]{dhall2012collecting}
Abhinav Dhall, Roland Goecke, Simon Lucey, Tom Gedeon, et~al.
\newblock Collecting large, richly annotated facial-expression databases from
  movies.
\newblock \emph{IEEE multimedia}, 19\penalty0 (3):\penalty0 34--41, 2012.

\bibitem[Ringeval et~al.(2013)Ringeval, Sonderegger, Sauer, and
  Lalanne]{ringeval2013introducing}
Fabien Ringeval, Andreas Sonderegger, Juergen Sauer, and Denis Lalanne.
\newblock Introducing the recola multimodal corpus of remote collaborative and
  affective interactions.
\newblock In \emph{2013 10th IEEE international conference and workshops on
  automatic face and gesture recognition (FG)}, pages 1--8. IEEE, 2013.

\bibitem[Bilakhia et~al.(2015)Bilakhia, Petridis, Nijholt, and
  Pantic]{bilakhia2015mahnob}
Sanjay Bilakhia, Stavros Petridis, Anton Nijholt, and Maja Pantic.
\newblock The mahnob mimicry database: A database of naturalistic human
  interactions.
\newblock \emph{Pattern recognition letters}, 66:\penalty0 52--61, 2015.

\bibitem[Wang et~al.(2017)Wang, Meghawat, Morency, and Xing]{wang2017select}
Haohan Wang, Aaksha Meghawat, Louis-Philippe Morency, and Eric~P Xing.
\newblock Select-additive learning: Improving generalization in multimodal
  sentiment analysis.
\newblock In \emph{2017 IEEE International Conference on Multimedia and Expo
  (ICME)}, pages 949--954. IEEE, 2017.

\bibitem[Busso et~al.(2008)Busso, Bulut, Lee, Kazemzadeh, Mower, Kim, Chang,
  Lee, and Narayanan]{busso2008iemocap}
Carlos Busso, Murtaza Bulut, Chi-Chun Lee, Abe Kazemzadeh, Emily Mower, Samuel
  Kim, Jeannette~N Chang, Sungbok Lee, and Shrikanth~S Narayanan.
\newblock Iemocap: Interactive emotional dyadic motion capture database.
\newblock \emph{Language resources and evaluation}, 42\penalty0 (4):\penalty0
  335, 2008.

\bibitem[Morency et~al.(2011)Morency, Mihalcea, and Doshi]{morency2011towards}
Louis-Philippe Morency, Rada Mihalcea, and Payal Doshi.
\newblock Towards multimodal sentiment analysis: Harvesting opinions from the
  web.
\newblock In \emph{Proceedings of the 13th international conference on
  multimodal interfaces}, pages 169--176. ACM, 2011.

\bibitem[W{\"o}llmer et~al.(2013)W{\"o}llmer, Weninger, Knaup, Schuller, Sun,
  Sagae, and Morency]{wollmer2013youtube}
Martin W{\"o}llmer, Felix Weninger, Tobias Knaup, Bj{\"o}rn Schuller, Congkai
  Sun, Kenji Sagae, and Louis-Philippe Morency.
\newblock Youtube movie reviews: Sentiment analysis in an audio-visual context.
\newblock \emph{IEEE Intelligent Systems}, 28\penalty0 (3):\penalty0 46--53,
  2013.

\bibitem[P{\'e}rez-Rosas et~al.(2013)P{\'e}rez-Rosas, Mihalcea, and
  Morency]{perez2013utterance}
Ver{\'o}nica P{\'e}rez-Rosas, Rada Mihalcea, and Louis-Philippe Morency.
\newblock Utterance-level multimodal sentiment analysis.
\newblock In \emph{Proceedings of the 51st Annual Meeting of the Association
  for Computational Linguistics (Volume 1: Long Papers)}, pages 973--982, 2013.

\bibitem[Poria et~al.(2018)Poria, Hazarika, Majumder, Naik, Cambria, and
  Mihalcea]{poria2018meld}
Soujanya Poria, Devamanyu Hazarika, Navonil Majumder, Gautam Naik, Erik
  Cambria, and Rada Mihalcea.
\newblock Meld: A multimodal multi-party dataset for emotion recognition in
  conversations.
\newblock \emph{arXiv preprint arXiv:1810.02508}, 2018.

\bibitem[Poria et~al.(2016)Poria, Chaturvedi, Cambria, and
  Hussain]{poria2016convolutional}
Soujanya Poria, Iti Chaturvedi, Erik Cambria, and Amir Hussain.
\newblock Convolutional mkl based multimodal emotion recognition and sentiment
  analysis.
\newblock In \emph{2016 IEEE 16th international conference on data mining
  (ICDM)}, pages 439--448. IEEE, 2016.

\bibitem[Poria et~al.(2015)Poria, Cambria, and Gelbukh]{poria2015deep}
Soujanya Poria, Erik Cambria, and Alexander Gelbukh.
\newblock Deep convolutional neural network textual features and multiple
  kernel learning for utterance-level multimodal sentiment analysis.
\newblock In \emph{Proceedings of the 2015 conference on empirical methods in
  natural language processing}, pages 2539--2544, 2015.

\bibitem[Xu et~al.(2013)Xu, Tao, and Xu]{xu2013survey}
Chang Xu, Dacheng Tao, and Chao Xu.
\newblock A survey on multi-view learning.
\newblock \emph{arXiv preprint arXiv:1304.5634}, 2013.

\bibitem[W{\"o}rtwein and Scherer(2017)]{wortwein2017really}
Torsten W{\"o}rtwein and Stefan Scherer.
\newblock What really matters—an information gain analysis of questions and
  reactions in automated ptsd screenings.
\newblock In \emph{2017 Seventh International Conference on Affective Computing
  and Intelligent Interaction (ACII)}, pages 15--20. IEEE, 2017.

\bibitem[Nojavanasghari et~al.(2016)Nojavanasghari, Gopinath, Koushik,
  Baltru{\v{s}}aitis, and Morency]{nojavanasghari2016deep}
Behnaz Nojavanasghari, Deepak Gopinath, Jayanth Koushik, Tadas
  Baltru{\v{s}}aitis, and Louis-Philippe Morency.
\newblock Deep multimodal fusion for persuasiveness prediction.
\newblock In \emph{Proceedings of the 18th ACM International Conference on
  Multimodal Interaction}, pages 284--288. ACM, 2016.

\bibitem[Pham et~al.(2018)Pham, Manzini, Liang, and
  Poczos]{pham2018seq2seq2sentiment}
Hai Pham, Thomas Manzini, Paul~Pu Liang, and Barnabas Poczos.
\newblock Seq2seq2sentiment: Multimodal sequence to sequence models for
  sentiment analysis.
\newblock \emph{arXiv preprint arXiv:1807.03915}, 2018.

\bibitem[Zadeh et~al.(2018{\natexlab{c}})Zadeh, Liang, Poria, Vij, Cambria, and
  Morency]{zadeh2018MARN}
Amir Zadeh, Paul~Pu Liang, Soujanya Poria, Prateek Vij, Erik Cambria, and
  Louis-Philippe Morency.
\newblock Multi-attention recurrent network for human communication
  comprehension.
\newblock In \emph{Thirty-Second AAAI Conference on Artificial Intelligence},
  2018{\natexlab{c}}.

\bibitem[Liang et~al.(2018)Liang, Liu, Zadeh, and Morency]{liang2018multimodal}
Paul~Pu Liang, Ziyin Liu, Amir Zadeh, and Louis-Philippe Morency.
\newblock Multimodal language analysis with recurrent multistage fusion.
\newblock \emph{arXiv preprint arXiv:1808.03920}, 2018.

\bibitem[Ekman et~al.(1980)Ekman, Freisen, and Ancoli]{ekman1980facial}
Paul Ekman, Wallace~V Freisen, and Sonia Ancoli.
\newblock Facial signs of emotional experience.
\newblock \emph{Journal of personality and social psychology}, 39\penalty0
  (6):\penalty0 1125, 1980.

\bibitem[Gu et~al.(2018{\natexlab{b}})Gu, Yang, Fu, Chen, Li, and
  Marsic]{gu2018multimodal}
Yue Gu, Kangning Yang, Shiyu Fu, Shuhong Chen, Xinyu Li, and Ivan Marsic.
\newblock Multimodal affective analysis using hierarchical attention strategy
  with word-level alignment.
\newblock \emph{arXiv preprint arXiv:1805.08660}, 2018{\natexlab{b}}.

\bibitem[Levinson and Holler(2014)]{levinson2014origin}
Stephen~C Levinson and Judith Holler.
\newblock The origin of human multi-modal communication.
\newblock \emph{Philosophical Transactions of the Royal Society B: Biological
  Sciences}, 369\penalty0 (1651):\penalty0 20130302, 2014.

\end{thebibliography}
\bibliographystyle{unsrtnat}
}

\section*{Supplementary}

\subsection*{Related Work}
Previous work on multimodal reasoning focuses on audio-visual data  \cite{grimm2008vera,dhall2012collecting,ringeval2013introducing,bilakhia2015mahnob,wang2017select}. Human communication uses both verbal and nonverbal modalities. Ignoring the contribution of text can cause misinterpretation as shown in example (b) in Figure \ref{fig:WhyMultiModal}. Recent work recognizes the importance of adding text as a modality, leading to a blossom of language-audio-visual datasets in tasks such as emotion recognition and sentiment analysis \cite{busso2008iemocap,morency2011towards,wollmer2013youtube,perez2013utterance,zadeh2016multimodal,zadeh2018multimodal,poria2018meld}. 
Models in previous work can be divided into three major categories: (1) early fusion, (2) late fusion, and (3) multi-view learning.

\textit{Early Fusion} concatenates input-level feature from different modalities. The fused features are then passed to a generic model. Intra-modal dynamics are not modeled explicitly when using this method of fusion \cite{poria2016convolutional,poria2015deep,wang2017select}.  Early fusion methods can preserve the temporal nature of data by using recurrent neural networks, or totally ignores the time factor when using classical machine learning methods such as Support Vector Machines \cite{zadeh2016multimodal,morency2011towards}. Due to the lack of specific intra-modal models, this fusion method tends to overfit on small datasets \cite{xu2013survey}.

\textit{Late Fusion} models each modality separately, then the outputs are combined in the decision phase. This can be done by majority voting , weighted averaging \cite{wortwein2017really}, or the output of modality networks are combined with a joint neural network that is trained to make finial decision \cite{nojavanasghari2016deep}.

\textit{Multi-view learning} is a broader class of methods that aim to learn both 
intra-modal dynamics and inter-modality dynamics jointly. To place our work in the context of prior research we divide multi-view learning into the following subcategories: \textbf{(a) Non-attention based fusion methods:} this includes tensor-based multimodal representations created from expensive tensor product \cite{zadeh2017tensor} or low-rank tensors approximation \cite{liu2018efficient}, generative representation learning \cite{pham2018seq2seq2sentiment,ngiam2011multimodal}, word-level factorized multimodal representation \cite{tsai2018learning} and variations to word representations by word-level fusion with nonverbal features \cite{wang2019words}. Such methods (excluding  \cite{liu2018efficient}) are rather slow and  suffer from exponentially increasing computational complexity as the number of modalities increase.
\textbf{(b) Attention based fusion methods:} The work in \cite{zadeh2018memory,zadeh2018MARN} used attention to discover interactions between modalities through time. Method proposed in \cite{liang2018multimodal} decomposes the fusion problem into multiple stages and focuses on a subset of multimodal signals at each stage.
In \cite{gu2018hybrid}, authors apply both feature attention and modality attention to classify utterance-level speech data.

MAN belongs to the attention based fusion subcategory; however, we differ from current methods by our ability to view modalities independently and quantize how each modality influences the final decision of the network. Additionally, the pre-training step allows our model to produce competitive results when compared to more complex architectures.

\subsection*{Detailed description of the experiments}
\paragraph{Datasets}
Details of each dataset is described in the section below and data split is shown in Table \ref{tab:DS_Stat}. Each datasets that contain spoken language, acoustic and visual information. 
\begin{table}[htbp!]
  \centering
    \begin{tabular}{l|c|c|c}
    \toprule
    Dataset & CMU-MOSI  & CMU-MOSEI  & POM \\
    \midrule
    \midrule
    \# Training & 1281  & 16265 & 560 \\
    \# Validation & 229   & 1869  & 92 \\
    \# Testing & 685   & 4643  & 188 \\
    \hline
    \end{tabular}%
    \caption{Number of training, validation and testing samples in each dataset.}
      \label{tab:DS_Stat}%
\end{table}%

\paragraph{Multimodal Sentiment Analysis:} \textbf{CMU-MOSI} dataset  \cite{zadeh2016multimodal} is a collection of 
93 review videos in English with 2199 utterance segments. Each utterance is annotated with sentiment by five individual annotators in the range [-3,3], where -3 indicates
highly negative and 3 indicates highly positive.
\paragraph{Multimodal Emotion Recognition:} \textbf{CMU-MOSEI}  dataset \cite{zadeh2018multimodal} is
a collection of 3,229 videos spanning over 23,000 utterances
from more than 1,000 online YouTube speakers. Each utterance is annotated with sentiment in the range [-3, 3] similar to CMU-MOSI dataset. In addition, utterances are annotated for Ekman emotions \cite{ekman1980facial} of {happiness, sadness, anger, fear, disgust, and surprise} with values in range [0, 3].
\paragraph{Multimodal Speaker Trait Recognition:} \textbf{POM}
Persuasion Opinion Multimodal dataset
\cite{park2014computational}  contains
1,000 movie review videos annotated for 16 different
speaker traits: confidence, passion, voice pleasant,  dominance, credibility, vividness, expertise, entertaining, reserved, trusting, relaxed, outgoing, thorough, nervous, humorous, and persuasive.

\paragraph{Feature Embedding}
Following prior practice \cite{liu2018efficient,liang2018multimodal,gu2018multimodal,wang2019words}, we use features provided by CMU-Multimodal SDK.\footnote{\url{https://github.com/A2Zadeh/CMU-MultimodalSDK}}  
\begin{itemize}
    \item \textbf{Language Features:} Pre-trained word embeddings GloVe \cite{pennington2014glove} are used to convert the transcripts of videos into sequence of 300-dimensional word embeddings.
        \item \textbf{Acoustic Features:} COVAREP  \cite{degottex2014covarep} acoustic analysis framework is used to extract low level acoustic features. The features include pitch tracking, polarity estimation, glottal closure instants, spectral envelope, glottal flow estimation and 12 Melfrequency cepstral coefficients along with other features.
    \item \textbf{Visual Features:} Facial expression analysis toolkit
FACET\footnote{\url{https:imotions.com}} is used as visual feature extractor. Features include facial action units, facial landmarks, head pose, gaze tracking and HOG features.
\end{itemize}

\subsubsection*{Baseline Models}

\indent\textit{Late Fusion:} A sub-network is learned for each modality such as language, acoustic, and visual. As a baseline, we use a bidirectional LSTM for each sub-network.  The last hidden layer from sub-networks are concatenated together and passed to a dense layer. This is referred to as \textbf{LF-LSTM}.

\indent\textit{Hybrid Attention based Multimodal Network \textbf{(MAF)}} \cite{gu2018hybrid}  a deep multimodal network with both
feature attention and modality attention. The hybrid
attention architecture helps the system focus on learning informative representations for both modality-specific feature extraction and model fusion.

\indent \textit{Tensor Fusion Network  \textbf{(TFN)}} \cite{zadeh2017tensor} learns intra-modality
and inter-modality dynamics by
creating a multi-dimensional tensor that captures unimodal,
bimodal and trimodal interactions across modalities.

\indent\textit{Low-rank Multimodal Fusion \textbf{(LMF)}} \cite{liu2018efficient}  performs multimodal fusion using low-rank tensors to improve efficiency.

\indent \textit{Memory Fusion Network \textbf{(MFN)}} \cite{zadeh2018memory} a multi-view sequential learning neural network  that  accounts for both view-specific and cross-view interactions through time. View-specific interactions are
learned through assigning an LSTM function to
each view. The cross-view interactions are identified using a special attention mechanism and summarized through time
with a multi-view gated Memory. 

\indent \textit{ Multimodal Factorization Model \textbf{(MFM)}} \cite{tsai2018learning} a model that factorizes representations into two sets of independent factors: multimodal discriminative
and modality-specific generative factors then optimizes for a joint generative-discriminative objective across multimodal data and labels.

\indent \textit{Recurrent Attended Variation Embedding Network \textbf{(RAVEN)}} \cite{wang2019words} a model for human multimodal language that considers the fine-grained structure of nonverbal subword sequences and dynamically shifts the
word representations based on these nonverbal cues.

\subsection*{Training Models}
All models are trained on  NVIDIA Tesla K80 GPU. For each dataset, we use the hyper-parameters reported by the baseline methods for that particular dataset. If the hyper-parameters were not reported in the paper, we perform an extensive grid search to find the best performing hyper-parameters. We use the same grid search method to find the best hyper-parameters for our method as well.

\subsection*{Understanding the Importance of Different Modalities through Attention}

\begin{figure}[hbt!]\centering
\subfloat[]{\label{a}\includegraphics[width=.4\linewidth]{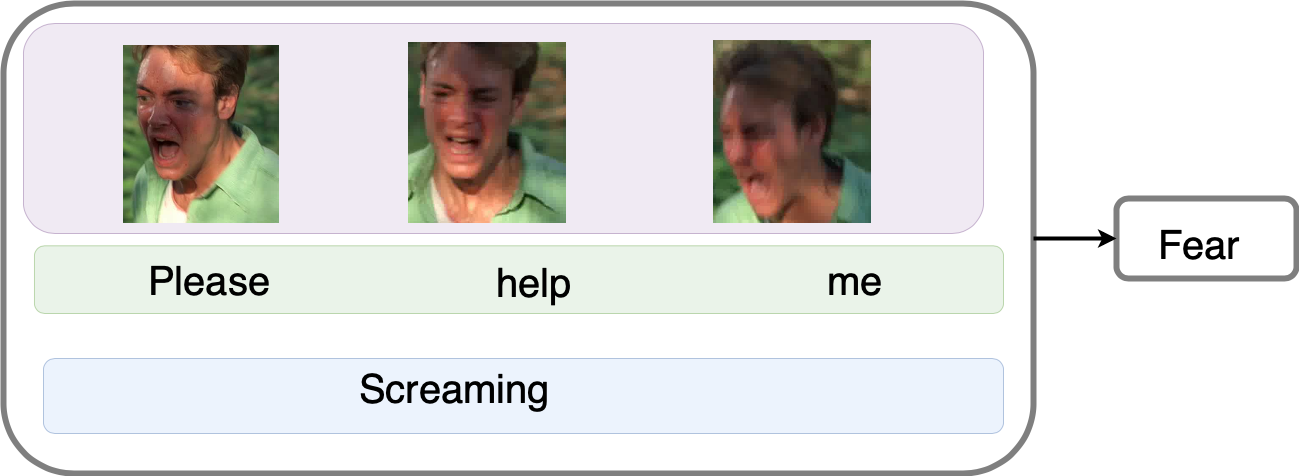}}\hfill
\subfloat[]{\label{b}\includegraphics[width=.4\linewidth]{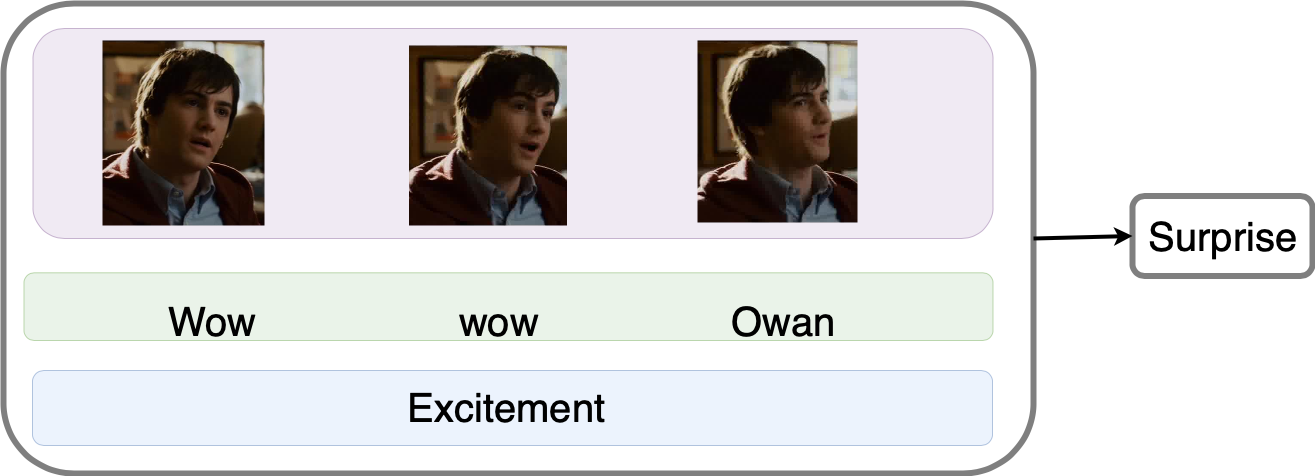}}\par 
\subfloat[]{\label{c}\includegraphics[width=.4\linewidth]{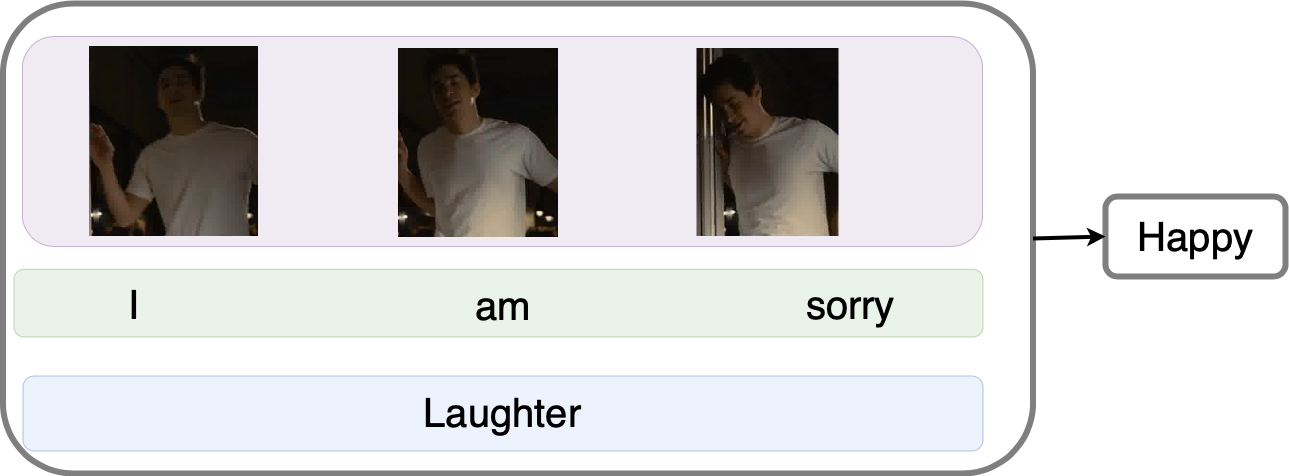} }
 \caption{Video examples for an emotion classification task. \textbf{(a)} Visual images are enough to make correct classification.  \textbf{(b)} Language is most informative modality.  \textbf{(c)} Visual and language modalities are misleading and acoustic features are most important.}%
    \label{fig:WhyMultiModal}%
\end{figure}

Human's ability to communicate effectively comes from their understanding of different modalities \cite{levinson2014origin}. The context in which a modality occurs changes its underlying meaning; a laugh can be both happy or sarcastic. In such cases, not all modalities are equally important for revealing the meaning of human interaction. 
Consider examples in Figure \ref{fig:WhyMultiModal} for an emotion classification task: In example (a), we can easily perceive that the emotion is fear as the visual signal is very prominent.
In example (b), visual signals are ambiguous and not useful; however, by looking at the sentence, the word "Wow" allows us to conclude that the emotion is surprise. In example (c), visual and language signals are misleading and acoustic features help make correct perception.  Humans can identify the important modality in cases where some modalities are ambiguous or may lead to wrong judgment.

\begin{figure}[hbt!]\centering
\subfloat[For this emotion recognition task in MOSEI, looking at any modality separately one can deduce that the speaker is happy; MAN gave modalities similar weights.]{\label{a }\includegraphics[width=1\linewidth]{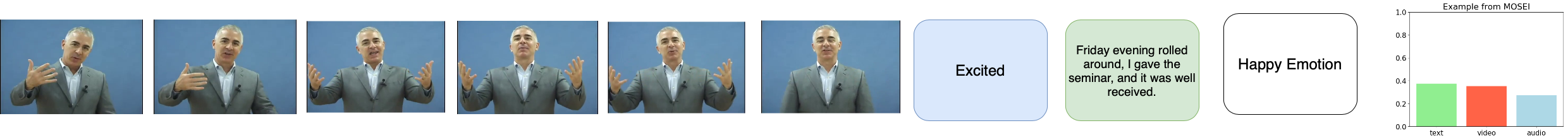}}\hfill
\subfloat[In a personality trait recognition task from POM dataset, the calmness in the speakers voice manifests how confident the speaker is, MAN assigns the highest weights to the acoustic modality.]{\label{b }\includegraphics[width=1\linewidth]{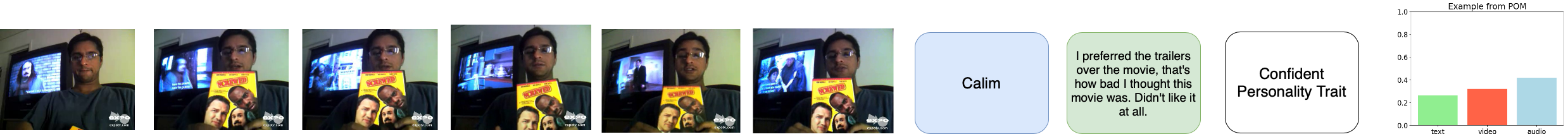}}
 \caption{Video examples from three datasets on different classification tasks; in each example we show the attention weight MAN produced for each modality. }%
    \label{fig:interpretation2}%
\end{figure}

The goal of the attention block in MAN is to mimic human's ability to understand the context of multimodal data by identifying the importance of modality. In Figure \ref{fig:interpretation}, we showed an example from MOSI where MAN can successfully detect the essence of different modalities. Figure \ref{fig:interpretation2} shows  examples from MOSEI and POM. Figure \ref{fig:interpretation2} (a) An example from MOSEI dataset on emotion classification task, all modalities point to happy emotion; MAN also produces similar weights for all modalities. Figure \ref{fig:interpretation2} (b) An example from POM dataset on personality trait recognition. Here the speaker is confident; it is difficult to recognize this from text but the calmness in the speaker's voice helped the model to make a correct prediction. MAN's ability to differentiate between modalities makes it in some sense more explainable.

\end{document}